\documentclass[11pt,a4paper]{article}
\usepackage[hyperref]{acl2021}
\usepackage{times}
\usepackage{latexsym}
\usepackage{graphicx}
\usepackage{amsthm,amsmath,amssymb}
\usepackage{multirow}
\usepackage{booktabs}
\usepackage{hyperref}[breaklinks]

\usepackage{microtype}

\aclfinalcopy 

\title{Marginal Utility Diminishes: Exploring the Minimum Knowledge for
BERT Knowledge Distillation}

\author{

    Yuanxin Liu\textsuperscript{\rm 1,2}\thanks{\quad Work was done when Yuanxin Liu was an intern at Pattern Recognition Center, WeChat AI, Tencent Inc, China.}, 
    Fandong Meng\textsuperscript{\rm 3},
    Zheng Lin\textsuperscript{\rm 1}\thanks{\quad Zheng Lin is the corresponding author.},
    Weiping Wang\textsuperscript{\rm 1},
    Jie Zhou\textsuperscript{\rm 3}
    \\
\textsuperscript{\rm 1}Institute of Information Engineering, Chinese Academy of Sciences, Beijing, China\\
\textsuperscript{\rm 2}School of Cyber Security, University of Chinese Academy of Sciences, Beijing, China \\
\textsuperscript{\rm 3}Pattern Recognition Center, WeChat AI, Tencent Inc, China \\
\{liuyuanxin,linzheng,wangweiping\}@iie.ac.cn, \{fandongmeng,withtomzhou\}@tencent.com
}

\date{}

\begin{document}
\maketitle
\begin{abstract}
Recently, knowledge distillation (KD) has shown great success in BERT compression. Instead of only learning from the teacher's soft label as in conventional KD, researchers find that the rich information contained in the hidden layers of BERT is conducive to the student's performance. To better exploit the hidden knowledge, a common practice is to force the student to deeply mimic the teacher's hidden states of all the tokens in a layer-wise manner. In this paper, however, we observe that although distilling the teacher's hidden state knowledge (HSK) is helpful, the performance gain (marginal utility) diminishes quickly as more HSK is distilled. To understand this effect, we conduct a series of analysis. Specifically, we divide the HSK of BERT into three dimensions, namely depth, length and width. We first investigate a variety of strategies to extract crucial knowledge for each single dimension and then jointly compress the three dimensions. In this way, we show that 1) the student's performance can be improved by extracting and distilling the crucial HSK, and 2) using a tiny fraction of HSK can achieve the same performance as extensive HSK distillation. Based on the second finding, we further propose an efficient KD paradigm to compress BERT, which does not require loading the teacher during the training of student. For two kinds of student models and computing devices, the proposed KD paradigm gives rise to training speedup of 2.7$\times$ $\sim$3.4$\times$.
\end{abstract}

\section{Introduction}
Since the launch of BERT \cite{BERT}, pre-trained language models (PLMs) have been advancing the state-of-the arts (SOTAs) in a wide range of NLP tasks. At the same time, the growing size of PLMs has inspired a wave of research interest in model compression \cite{Song16ICLR} in the NLP community, which aims to facilitate the deployment of the powerful PLMs to resource-limited scenarios. 

Knowledge distillation (KD) \cite{HintonKD} is an effective technique in model compression. In conventional KD, the student model is trained to imitate the teacher's prediction over classes, i.e., the soft labels. Subsequently, \citet{FitNets} find that the intermediate representations in the teacher's hidden layers can also serve as a useful source of knowledge. As an initial attempt to introduce this idea to BERT compression, PKD \cite{PKD} proposed to distill representations of the $[\mathrm{CLS}]$ token in BERT's hidden layers, and later studies \cite{TinyBERT,MobileBERT,DynaBERT,Rosita} extend the distillation of hidden state knowledge (HSK) to all the tokens.

\begin{figure}[t]
\centering
\includegraphics[width=0.8\linewidth]{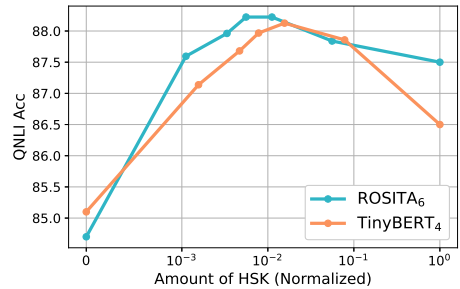}
\caption{The Acc variation of ROSITA \cite{Rosita} and TinyBERT \cite{TinyBERT} on QNLI with the increase of HSK.}
\label{fig:marginal-utility-qnli}
\end{figure}

In contrast to the previous work that attempts to increase the amount of HSK, in this paper we explore towards the opposite direction to ``compress" HSK. We make the observation that although distilling HSK is helpful, the marginal utility diminishes quickly as the amount of HSK increases. To understand this effect, we conduct a series of analysis by compressing the HSK from three dimensions, namely depth, length and width (see Section \ref{sec:hsk_compress} for detailed description). We first compress each single dimension and compare a variety of strategies to extract crucial knowledge. Then, we jointly compress the three dimensions using a set of compression configurations, which specify the amount of HSK assigned to each dimension. Figure \ref{fig:marginal-utility-qnli} shows the results on QNLI dataset. We can find that 1) perceivable performance improvement can be obtained by extracting and distilling the crucial HSK, and 2) with only a tiny fraction of HSK the students can achieve the same performance as extensive HSK distillation.

Based on the second finding, we further propose an efficient paradigm to distill HSK. Concretely, we run BERT over the training set to obtain and store a subset of HSK. This can be done on cloud devices with sufficient computational capability. Given a target device with limited resource, we can compress BERT and select the amount of HSK accordingly. Then, the compressed model can perform KD on either the cloud or directly on the target device using the selected HSK and the original training data, dispensing with the need to load the teacher model. 

In summary, our maojor contributions are:
\begin{itemize}
\item We observe the marginal utility diminishing effect of HSK in BERT KD. To our knowledge, we are the first attempt to systematically study knowledge compression in BERT KD.
\item We conduct exploratory studies on how to extract the crucial knowledge in HSK, based on which we obtain perceivable improvements over a widely-used HSK distillation strategy.
\item We propose an efficient KD paradigm based on the empirical findings. Experiments on the GLUE benchmark for NLU \cite{GLUE} show that, the proposal gives rise to training speedup of 2.7$\times$ $\sim$3.4$\times$ for TinyBERT and ROSITA on GPU and CPU\footnote{The code will be available at https://github.com/llyx97/Marginal-Utility-Diminishes}.
\end{itemize}

\section{Preliminaries}

\subsection{BERT Architecture}
The backbone of BERT consists of an embedding layer and $L$ identical Transformer \cite{Transformer} layers. The input to the embedding layer is a text sequence $\mathbf{x}$ tokenized by WordPiece \cite{WordPiece}. There are two special tokens in $\mathbf{x}$: $[\mathrm{CLS}]$ is inserted in the left-most position to aggregate the sequence representation and $[\mathrm{SEP}]$ is used to separate text segments. By summing up the token embedding, the position embedding and the segment embedding, the embedding layer outputs a sequence of vectors $\mathbf{E} = \left[\mathbf{e}_{1}, \cdots, \mathbf{e}_{|\mathbf{x}|}\right] \in \mathbb{R}^{|\mathbf{x}| \times d_{H}}$, where $d_{H}$ is the hidden size of the model.

Then, $\mathbf{E}$ passes through the stacked Transformer layers, which can be formulated as:
\begin{equation}
\mathbf{H}_{l}=\operatorname{Trm}_{l}\left(\mathbf{H}_{l-1}\right), l \in[1, L]
\end{equation}
where $\mathbf{H}_{l}=\left[\mathbf{h}_{l,1}, \cdots, \mathbf{h}_{l,|\mathbf{x}|}\right] \in \mathbb{R}^{|\mathbf{x}| \times d_{H}}$ is the outputs of the $l^{th}$ layer and $\mathbf{H}_{0}=\mathbf{E}$. Each Transformer layer is composed of two sub-layers: the multi-head self-attention layer and the feed-forward network (FFN). Each sub-layer is followed by a sequence of dropout \cite{Dropout}, residual connection \cite{ResNet} and layer normalization \cite{layer-norm}.

Finally, for the tasks of NLU, a task-specific classifier is employed by taking as input the representation of $[\mathrm{CLS}]$ in  the $L^{th}$ layer.

\subsection{BERT Compression with KD}
Knowledge distillation is a widely-used technique in model compression, where the compressed model (student) is trained under the guidance of the original model (teacher). This is achieved by minimizing the difference between the features produced by the teacher and the student:
\begin{equation}
\mathcal{L}_{\mathrm{KD}}=\sum_{\left(f^{S}, f^{T}\right)} \mathcal{L}\left(f^{S}(\mathbf{x}), f^{T}(\mathbf{x})\right)
\end{equation}
where $\left(f^{S}, f^{T}\right)$ is a pair of features from student and teacher respectively. $\mathcal{L}$ is the loss function and $\mathbf{x}$ is a data sample. In terms of BERT compression, the predicted probability over classes, the intermediate representations and the self-attention distributions can be used as the features to transfer. In this paper, we focus on the intermediate representations $\left\{\mathbf{H}_{l}\right\}_{l=0}^{L}$ (i.e., the HSK), which have shown to be a useful source of knowledge in BERT compression. The loss function is computed as the Mean Squared Error (MSE) in a layer-wise way:
\begin{equation}
\mathcal{L}_{HSK}=\sum_{l=0}^{L^{'}} \operatorname{MSE}\left(\mathbf{H}^{S}_{l} \mathbf{W},\mathbf{H}^{T}_{g(l)}\right)
\label{eq:hsk_distill}
\end{equation}
where $L^{'}$ is the student's layer number and $g(l)$ is the layer mapping function to select teacher layers. $\mathbf{W} \in \mathbb{R}^{d^{S}_{H} \times d^{T}_{H}}$ is the linear transformation to project the student's representations $\mathbf{H}^{S}_{l}$ to the same size as the teacher's representation $\mathbf{H}^{T}_{l}$.

\subsection{HSK Compression}
\label{sec:hsk_compress}
According to Equation \ref{eq:hsk_distill}, the HSK from teacher can be stacked into a tensor $\mathbf{\widehat{H}}^{T}=\left[\mathbf{H}^{T}_{g(0)}, \cdots, \mathbf{H}^{T}_{g(L^{'})}\right] \in \mathbb{R}^{(L^{'}+1) \times |\mathbf{x}| \times d^{T}_{H}}$, which consists of three structural dimensions, namely depth, length and width. For the depth dimension, $\mathbf{\widehat{H}}^{T}$ can be compressed by eliminating entire layers. By dropping the representations corresponding to particular tokens, we compress the length dimension. When it comes to the width dimension, we set the eliminated activations to zero. We will discuss the strategies to compress each dimension later in Section~\ref{sec:single_dim}.

\section{Experimental Setups}
\subsection{Datasets}
We perform experiments on seven tasks from the General Language Understanding Evaluation (GLUE) benchmark \cite{GLUE}: CoLA (linguistic acceptability), SST-2 (sentiment analysis), RTE, QNLI, MNLI-m and MNLI-mm (natural language inference), MRPC and STS-B (semantic matching/similarity). Due to space limitation, we only report results on CoLA, SST-2, QNLI and MNLI for single-dimension HSK compression in Section \ref{sec:single_dim}, and results on the other three tasks are presented in Appendix \ref{sec:appendix_more_results}.

\subsection{Evaluation}
Following \citep{BERT}, for the dev set, we use Matthew’s correlation and Spearman correlation to evaluate the performance on CoLA and STS-B respectively. For the other tasks, we report the classification accuracy. We use the dev set to conduct our exploratory studies and the test set results are reported to compare HSK compression with the existing distillation strategy. For the test set of MRPC, we report the results of F1 score.

\subsection{Implementation Details}
We take two representative KD-based methods, i.e., TinyBERT \cite{TinyBERT} and ROSITA \cite{Rosita}, as examples to conduct our analysis. TinyBERT is a compact version of BERT that is randomly initialized. It is trained with two-stage KD: first on the unlabeled general domain data and then on the task-specific training data. ROSITA replaces the first stage KD with structured pruning and matrix factorization, which can be seen as a direct transfer of BERT's knowledge from the model parameters. 

We focus on KD with the task-specific training data and do not use any data augmentation. For TinyBERT, the student model is initialized with the 4-layer general distillation model provided by \citet{TinyBERT}
(denoted as $\text{TinyBERT}_{4}$). For ROSITA, we first fine-tune $\mathrm{BERT}_{\mathrm{BASE}}$ on the downstream task and then compress it following \citet{Rosita} to obtain a 6-layer student model (denoted as $\text{ROSITA}_{6}$). The fine-tuned $\mathrm{BERT}_{\mathrm{BASE}}$ is used as the shared teacher for TinyBERT and ROSITA. Following \citet{TinyBERT}, we first conduct HSK distillation as in Equation \ref{eq:hsk_distill} (w/o distilling the self-attention distribution) and then distill the teacher's predictions using cross-entropy loss. All the results are averaged over three runs with different random seeds. The model architecture of the students and the hyperparameter settings can be seen in Appendix \ref{sec:model_archi} and Appendix \ref{sec:hyperparameter} respectively.

\section{Single-Dimension Knowledge Compression}
\label{sec:single_dim}
Researches on model pruning have shown that the structural units in a model are of different levels of importance, and the unimportant ones can be dropped without affecting the performance. In this section, we investigate whether the same law holds for HSK compression in KD. We study the three dimensions separately and compare a variety of strategies to extract the crucial knowledge. When a certain dimension is compressed, the other two dimensions are kept to full scale.

\subsection{Depth Compression}
\subsubsection{Compression Strategies}
From the layer point of view, HSK compression can be divided into two steps. First, the layer mapping function $g(l)$ selects one of the teacher layers for each student layer. This produces $L^{'}+1$ pairs of teacher-student features: $\left[(\mathbf{H}^{S}_{0},\mathbf{H}^{T}_{g(0)}), \cdots, (\mathbf{H}^{S}_{L^{'}},\mathbf{H}^{T}_{g(L^{'})})\right]$. Second, a subset of these feature pairs are selected to perform HSK distillation. 

\begin{figure}[t]
\centering
\includegraphics[width=0.9\linewidth]{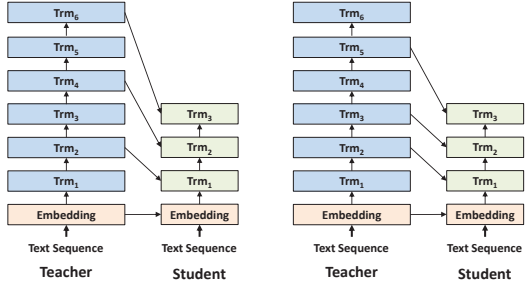}
\caption{Illustration of the redesigned uniform layer mapping strategy. Left: $L^{top}$ is the top teacher layer. Right: $L^{top}$ is the second-top teacher layer.}
\label{fig:layer-map}
\end{figure}

For the first step, a simple but effective strategy is the uniform mapping function:
\begin{equation}
    g(l)=l \times \frac{L}{L^{'}}, \bmod (L, L^{'})= 0
    \label{eq:uniform_mapping}
\end{equation}
In this way, the teacher layers are divided into $L^{'}$ blocks and the top layer of each block serves as the guidance in KD. Recently, \citet{MiniLMv2} empirically show that the upper-middle layers of BERT, as compared with the top layer, are a better choice to guide the top layer of student in self-attention distillation. Inspired by this, we redesign Equation \ref{eq:uniform_mapping} to allow the top student layer to distill knowledge from an upper-middle teacher layer, and the lower layers follow the uniform mapping principle. This function can be formulated as:
\begin{equation}
    g(l, L^{top})=l \times \operatorname{round}(\frac{L^{top}}{L^{'}})
    \label{eq:uniform_mapping_redesign}
\end{equation}
where $L^{top}$ is the teacher layer corresponding to the top student layer and $\operatorname{round}()$ is the rounding-off operation. Figure \ref{fig:layer-map} gives an illustration of $g(l, L^{top})$ with a 6-layer teacher and a 3-layer student. Specifically, for the 12-layer $\mathrm{BERT}_{\mathrm{BASE}}$ teacher, we select $L^{top}$ from $\{8, 10, 12\}$. For the second step, we simply keep the top $N^{D}$ feature pairs:
$\{(\mathbf{H}^{S}_{l},\mathbf{H}^{T}_{g(l,L^{top})})\}_{l=L^{'}-N^{D}+1}^{L^{'}}$.

\subsubsection{Results and Analysis}
\begin{figure}[t]
\centering
\includegraphics[width=0.9\linewidth]{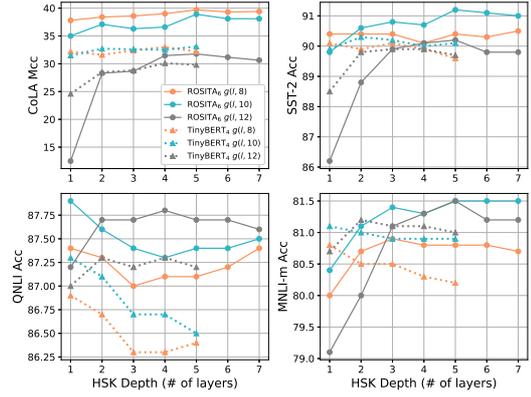}
\caption{Results of depth compression on CoLA, SST-2, QNLI and MNLI. Each color denotes a layer mapping function. The number of layers in HSK includes the embedding layer. Full results on seven tasks are shown in Appendix \ref{sec:appendix_depth}.}
\label{fig:depth-compress}
\end{figure}

Figure \ref{fig:depth-compress} presents the results of depth compression with different layer mapping functions. We can find that: 1) For the $g(l,12)$ mapping function (the grey lines), depth compression generally has a negative impact on the students' performance. Specially, the performance of $\text{ROSITA}_{6}$ declines drastically when the number of layers is reduced to $1 \sim 3$. 2) In terms of the $g(l,10)$ and $g(l,8)$ mapping functions (the blue and orange lines), HSK distillation with only one or two layers can achieve comparable performance as using all the $L^{'}+1$ layers. On the QNLI and MNLI datasets, the performance can even be improved by eliminating the lower layers. 3) In general, the student achieves better results with the redesigned layer mapping function in Equation \ref{eq:uniform_mapping_redesign} across the four tasks. This demonstrates that, like the self-attention knowledge, the most crucial HSK does not necessarily reside in the top BERT layer, which reveals a potential way to improve HSK distillation of BERT. 4) Compared with $g(l,8)$, the improvement brought by $g(l,10)$ is more stable across different tasks and student models. Therefore, we use the $g(l,10)$ layer mapping function when investigating the other two dimensions.

\subsection{Length Compression}
\subsubsection{Compression Strategies}
\label{sec:length_strategy}
To compress the length dimension, we design a method to measure the tokens' importance by using the teacher's self-attention distribution. The intuition is that self-attention controls the information flow among tokens across layers, and thus the representations of the most attended tokens may contain crucial information.

Assuming that the teacher has $A_{h}$ attention heads, and the attention weights in the $l^{th}$ layer is $\mathbf{A}^{T}_{l} = \left\{\mathbf{A}^{T}_{l,a}\right\}_{a=1}^{A_{h}}$, where $\mathbf{A}^{T}_{l,a} \in \mathbb{R}^{|\mathbf{x}| \times |\mathbf{x}|}$ is the attention matrix of the $a^{th}$ head. Each row of $\mathbf{A}^{T}_{l,a}$ is the attention distribution of a particular token to all the tokens. In our length compression strategy, the importance score of the tokens is the attention distribution of the $[\mathrm{CLS}]$ token (i.e., the first row in $\mathbf{A}^{T}_{l,a}$) averaged over the $A_{h}$ heads:
\begin{equation}
\mathbf{S}_{l}=\frac{1}{A_{h}} \sum_{a=1}^{A_{h}} \mathbf{A}_{l, a, 1}^{T}, \mathbf{S}_{l} \in \mathbb{R}^{|\mathbf{x}|}
\label{eq:importance_score}
\end{equation}
To match the depth of the student, we employ the layer mapping function in Equation \ref{eq:uniform_mapping_redesign} to select $\mathbf{S}_{g(l, L^{top})}$ for the $l^{th}$ student layer. 

The length compression strategies examined in this section are summarized as:

\paragraph{Att} is the attention-based strategy as described above. The layer mapping function to select $\mathbf{S}$ is the same as the one to select HSK, i.e., $g(l, 10)$.
\paragraph{Att w/o $[\mathrm{SEP}]$} excludes the HSK of the special token $[\mathrm{SEP}]$. The rationality of this operation will be explained in the following analysis.
\paragraph{Att ($L^{top}=12$) w/o $[\mathrm{SEP}]$} is different from \textbf{Att w/o $[\mathrm{SEP}]$} in that it utilizes $g(l, 12)$ to select $\mathbf{S}$.
\paragraph{Left} is a naive baseline that discards tokens from the tail of the text sequence. When the token number is reduced to 1, the student only distills the HSK from the $[\mathrm{CLS}]$ token.

\begin{figure}[t]
\centering
\includegraphics[width=0.9\linewidth]{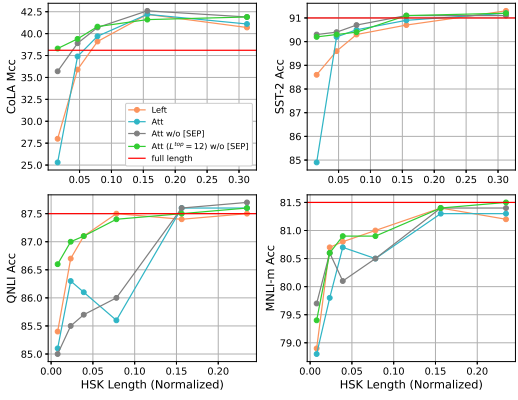}
\caption{Length compression results of $\text{ROSITA}_{6}$ on CoLA, SST-2, QNLI and MNLI. The horizontal axis represents the compressed HSK length normalized by full length. The left-most points in each plot mean compressing the length to one token. Full results on seven tasks are shown in Appendix \ref{sec:appendix_length}.}
\label{fig:length-rosita}
\end{figure}

\subsubsection{Results and Analysis}
\label{sec:length_compress_analy}
\begin{figure}[t]
\centering
\includegraphics[width=0.9\linewidth]{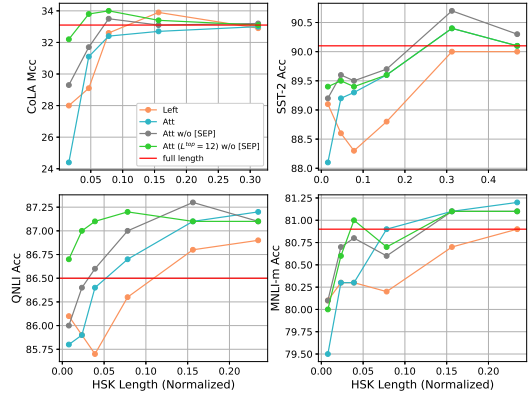}
\caption{Length compression results of $\text{TinyBERT}_{4}$ on CoLA, SST-2, QNLI and MNLI. The axes, points and lines are defined in the same way as Figure \ref{fig:length-rosita}. Full results on seven tasks are shown in Appendix \ref{sec:appendix_length}.}
\label{fig:length_tinybert}
\end{figure}

The length compression results are shown in Figure \ref{fig:length-rosita} and Figure \ref{fig:length_tinybert}. We can derive the following observations: 1) For all strategies, significant performance decline can only be observed when HSK length is compressed heavily (to less than $0.05 \sim 0.30$). In some cases, using a subset of tokens' representation even leads to perceivable improvement over the full length (e.g., $\text{ROSITA}_{6}$ on CoLA and $\text{TinyBERT}_{4}$ on SST-2 and QNLI). 2) The performance of \textbf{Att} is not satisfactory. When being applied to $\text{ROSITA}_{6}$, the \textbf{Att} strategy underperforms the \textbf{Left} baseline. The results of \textbf{Att} in $\text{TinyBERT}_{4}$, though better than those in $\text{ROSITA}_{6}$, still lag behind the other strategies at the left-most points. 3) Excluding $[\mathrm{SEP}]$ in the \textbf{Att} strategy alleviates the drop in performance, especially when HSK length is compressed to less than 0.05. 4) As a general trend, further improvement over \textbf{Att w/o $[\mathrm{SEP}]$} can be obtained by using $g(l, 12)$ in the selection of $\mathbf{S}$, which produces the most robust results among the four strategies.
\begin{figure}[t]
\centering
\includegraphics[width=1.0\linewidth]{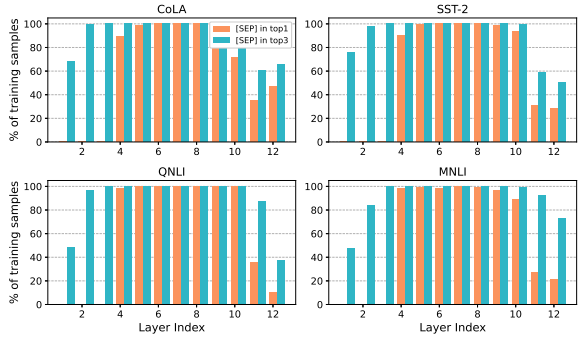}
\caption{The proportion of the data samples in which $[\mathrm{SEP}]$ is among the top1 and top3 attended tokens. We present the results over the 12 layers of the $\mathrm{BERT}_{\mathrm{BASE}}$ fine-tuned on CoLA, SST-2, QNLI and MNLI. Full results on seven tasks are shown in Appendix \ref{sec:appendix_length}.}
\label{fig:sep-rank}
\end{figure}

To explain why the \textbf{Att} strategy performs poorly, we inspect into the tokens that receive the highest importance scores under Equation \ref{eq:importance_score}. We find that the special token $[\mathrm{SEP}]$ is dominant in most hidden layers. As shown in Figure \ref{fig:sep-rank}, from the $4^{th} \sim 10^{th}$ layers, $[\mathrm{SEP}]$ is the most attended token for almost all training samples. Meanwhile, $[\mathrm{SEP}]$ frequently appears in the top three positions across all the layers. Similar phenomenon was found in \citet{clark2019what}, where $[\mathrm{SEP}]$ receives high attention scores from itself and other tokens in the middle layers. Combining this phenomenon and the results in Figure \ref{fig:length-rosita} and Figure \ref{fig:length_tinybert}, it can be inferred that the representations of $[\mathrm{SEP}]$ is not a desirable source of knowledge for ROSITA and TinyBERT. We conjecture that this is because there exists some trivial patterns in the representations of $[\mathrm{SEP}]$, which prevents the student to extract the informative features that are more relevant to the task.

\begin{figure}[t]
\centering
\includegraphics[width=0.9\linewidth]{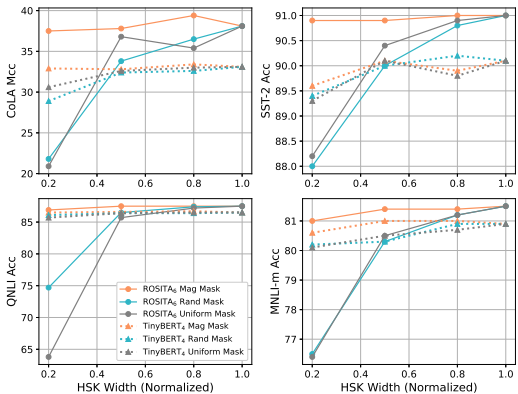}
\caption{Results of width compression with different masking strategies on CoLA, SST-2, QNLI and MNLI. Full results on seven tasks are shown in Appendix \ref{sec:appendix_width}.}
\label{fig:width}
\end{figure}

\subsection{Width Compression}
\subsubsection{Compression Strategies}
As discussed in Section \ref{sec:hsk_compress}, the width dimension is compressed by setting some activations in the intermediate representations to zero. Practically, we apply a binary mask $\mathbf{M} \in \mathbb{R}^{d^{T}_{H}}$ to the vectors in $\mathbf{H}^{T}_{l}$, which gives rise to $\left[\mathbf{M} \odot \mathbf{h}^{T}_{l,1}, \cdots, \mathbf{M} \odot \mathbf{h}^{T}_{l,|\mathbf{x}|}\right]$, where $\odot$ denotes the element-wise product. On this basis, we introduce and compare three masking designs for width compression:

\paragraph{Rand Mask} randomly set the values in $\mathbf{M}$ to zero, where the total number of ``0"  is controlled by the compression ratio. This mask is static, i.e., $\mathbf{h}^{T}_{l,i} (\forall i,l)$ for all the training samples share the same mask.
\paragraph{Uniform Mask} is also a static mask. It is constructed by distributing ``0" in a uniform way. Formally, the mask $\mathbf{M}$ is defined as:
\begin{equation}
\mathbf{M}_{i}=\left\{\begin{array}{cc}
1, & i \in \mathbf{I} \\
0, & \text { otherwise }
\end{array}\right.
\end{equation}
where $\mathbf{I}=\left\{\operatorname{round}\left(i \times \frac{d_{H}^{T}}{N^{W}}\right)\right\}_{i=1}^{N^{W}}$ is the indices of the remained $N^{W}$ activations.
\paragraph{Mag Mask} masks out the activations with low magnitude. Therefore, this mask is dynamic, i.e., every $\mathbf{h}^{T}_{l,i} (\forall i,l)$ has its own $\mathbf{M}$.

\subsubsection{Results and Analysis}
The width compression results can be seen in Figure \ref{fig:width}, from which we can obtain two findings. First, the masks reveal different patterns when combined with different student models. For $\text{ROSITA}_{6}$, the performance of \textbf{Rand Mask} and \textbf{Uniform Mask} decreases sharply at 20\% HSK width. In comparison, the performance change is not that significant when it comes to $\text{TinyBERT}_{4}$. This suggests that $\text{TinyBERT}_{4}$ is more robust to HSK width compression than $\text{ROSITA}_{6}$. Second, the magnitude-based masking strategy obviously outperforms \textbf{Rand Mask} and \textbf{Uniform Mask}. As we compress the nonzero activations in HSK from $100\%$ to $20\%$, the performance drop of \textbf{Mag Mask} is only marginal, indicating that there exists considerable knowledge redundancy in the width dimension.

\section{Three-Dimension Joint Knowledge Compression}
\label{sec:three_dim}
With the findings in single-dimension compression, we are now at a position to investigate joint HSK compression from the three dimensions.

\subsection{Measuring the Amount of HSK}
For every single dimension, measuring the amount of HSK is straightforward: using the number of layers, tokens and activations for depth, length and width respectively. In order to quantify the total amount of HSK (denoted as $A^{HSK}$), we define \textbf{one unit of $A^{HSK}$} as the amount of HSK in any $\mathbf{h}^{T}_{l,i} (\forall l \in [0,L],i \in [1, |\mathbf{x}|])$. In other words, the $A^{HSK}$ of $\mathbf{\widehat{H}}^{T}$ equals to $(L^{'}+1) \times |\mathbf{x}|$. When HSK is compressed to $N^{D}$ layers, $N^{L}$ tokens and $N^{W}$ activations, the $A^{HSK}$ is $N^{D} \times N^{L} \times \frac{N^{W}}{d_{H}^{T}}$.

\begin{table}[t]
\footnotesize
\centering
\resizebox{1.0\hsize}{!}{$
\begin{tabular}{@{}l c c c c c c @{}}
\toprule
$A^{HSK}$/unit            &1$\pm10\%$    &3$\pm5\%$     &5$\pm5\%$       &10$\pm5\%$    &50$\pm5\%$ \\\midrule
$\text{ROSITA}_{6}$     &13   &20    &31      &36    &21  \\
$\text{TinyBERT}_{4}$   &13   &18    &26      &30    &13  \\
\bottomrule
\end{tabular}
$}
\caption{The number of sampled configurations for different $A^{HSK}$. Each $A^{HSK}$ is extended to $\pm5\%$ or $\pm10\%$ to include more configurations.}
\label{tab:3d_config}
\end{table}
\begin{figure*}[t]
\centering
\includegraphics[width=0.9\linewidth]{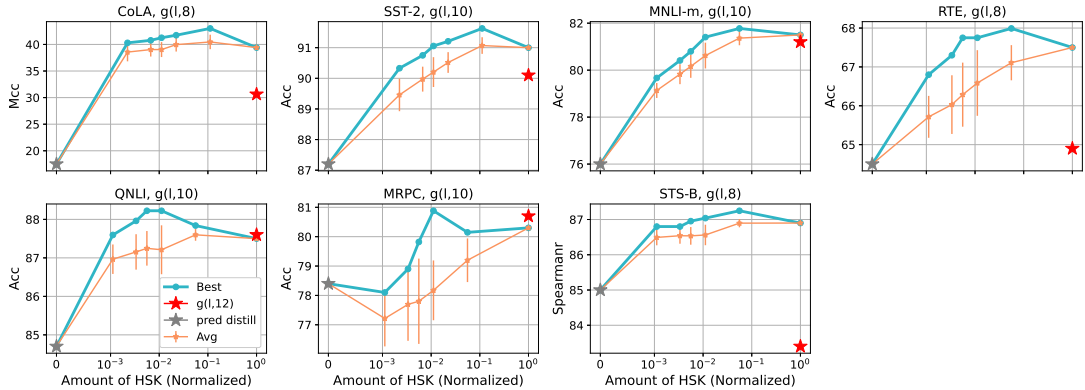}
\caption{Results of 3D HSK compression for $\text{ROSITA}_{6}$. The horizontal axis represents the remained $A^{HSK}$ normalized by the $A^{HSK}$ of $\mathbf{\widehat{H}}^{T}$. The left-most points (grey stars) in each plot correspond to only distilling the teacher's predictions. The mapping function used in depth compression is shown in the title of each plot, and the red stars denote the results of using $g(l,12)$. We show the averaged and best results of the configurations with the same $A^{HSK}$. The error bars of ``Avg" denote the standard deviation.}
\label{fig:3d_compress_rosita}
\end{figure*}
\begin{figure*}[t]
\centering
\includegraphics[width=0.9\linewidth]{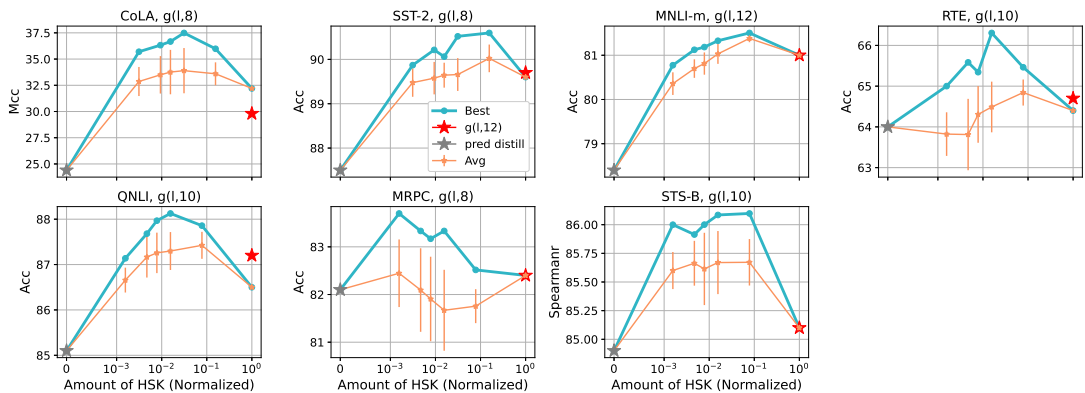}
\caption{Results of 3D HSK compression for $\text{TinyBERT}_{4}$. The axes, points, titles and lines are defined in the same way as Figure \ref{fig:3d_compress_rosita}.}
\label{fig:3d_compress_tinybert}
\end{figure*}
\subsection{Compression Configurations \& Strategies}
Formally, the triplet $(N^{D}, N^{L}, N^{W})$ defines a search space $\in \mathbb{R}^{(L^{'}+1) \times |\mathbf{x}| \times d^{T}_{H}}$ of the configurations for three-dimension (3D) HSK compression, and we could have multiple combinations of $(N^{D}, N^{L}, N^{W})$ that satisfy a particular $A^{HSK}$. In practice, we reconstruct the search space as:

\begin{equation}
\begin{array}{c}
\resizebox{1.0\hsize}{!}{$
\quad N^{D} \in [1, L^{'}+1], N^{L} \in [1, 50], \frac{N^{W}}{d_{H}^{T}} \in \{0.1\times i\}_{i=1}^{10}
$}
\end{array}
\end{equation}
To study the student's performance with different amounts of HSK, we sample a set of configurations for a range of $A^{HSK}$, the statistics of which is summarized in Table \ref{tab:3d_config}. Details of the configurations can be seen in Appendix \ref{sec:config}.

To compress each single dimension in joint HSK compression, we utilize the most advantageous strategies that we found in Section \ref{sec:single_dim}. Specifically, \textbf{Att ($L^{top}=12$) w/o $[\mathrm{SEP}]$} is used to compress length, \textbf{Mag Mask} is used to compress width and the $g(l,L^{top})$ for depth compression is selected according to the performance of depth compression.

\begin{table*}[t]
\small
\centering
\begin{tabular}{@{}l l c c c c c c c c@{}}
\toprule
&Method   &CoLA &SST-2 &QNLI &MNLI-m/mm  &MRPC  &RTE  &STS-B  &Avg\\ \midrule

\multirow{5}{*}{Dev}
& $\mathrm{BERT}_{\mathrm{BASE}}$(T)
&60.1  &93.5  &91.5    &84.7/84.7 &86.0  &67.5  &88.5  &82.1 \\ \cmidrule{2-10}

&$\text{TinyBERT}_{4}$    
&29.8   &89.7   &87.2   &81.0/81.4   &82.4   &64.7   &85.1   &75.2     \\
&\quad w/ HSK compression
&37.5   &90.6   &88.1   &81.5/81.7   &83.3   &66.3   &86.1   &76.9    \\

&$\text{ROSITA}_{6}$
&30.6   &90.1   &87.6   &81.2/81.5   &80.7   &64.9   &83.4   &75.0    \\

&\quad w/ HSK compression
&43.0   &91.6   &88.2   &81.8/82.0   &80.9   &68.0   &87.2   &77.8  \\
\midrule

\multirow{5}{*}{Test} 
& $\mathrm{BERT}_{\mathrm{BASE}}$(G)
&52.1   &93.5   &90.5   &84.6/83.4   &88.9  &66.4    &85.8  &80.7  \\ \cmidrule{2-10}

&$\text{TinyBERT}_{4}$ 
&28.2   &90.9   &86.4   &81.0/80.3   &85.6  &61.5    &76.8   &73.8   \\

&\quad w/ HSK compression
&30.6   &90.6   &87.3   &81.5/80.8   &85.4  &61.7    &79.0   &74.6   \\

&$\text{ROSITA}_{6}$      
&28.1   &90.5   &87.0   &81.5/80.4   &83.0  &61.7    &73.9   &73.3  \\

&\quad w/ HSK compression
&35.3   &91.3   &86.7   &81.9/80.9   &84.5  &61.7    &79.9   &75.3 \\

\bottomrule
\end{tabular}
\caption{Dev and test set performance of $\mathrm{BERT}_{\mathrm{BASE}}$ and KD-based BERT compression methods. (G) and (T) denote the results of $\mathrm{BERT}_{\mathrm{BASE}}$ from \citet{BERT} and the results of our teacher model, respectively.}
\label{tab:bert-compress}
\end{table*}

\subsection{Results and Analysis}
The results of 3D joint HSK compression are presented in Figure \ref{fig:3d_compress_rosita} and Figure \ref{fig:3d_compress_tinybert}. As we can see, introducing HSK in KD brings consistent improvement to the conventional prediction distillation method. However, the marginal benefit quickly diminishes as more HSK is included. Typically, with less than $1\%$ of HSK, the student models can achieve the same or better result as full-scale HSK distillation. Over a certain threshold of $A^{HSK}$, the performance begins to decrease. Among different tasks and student models, the gap between the best results (peaks on the blue lines) and full-scale HSK distillation varies from 0.3 ($\text{ROSITA}_{6}$ on MNLI and STS-B) to 5.3 ($\text{TinyBERT}_{4}$ on CoLA). The results also suggest that existing BERT distillation method (i.e., $g(l,12)$) can be improved by simply compressing HSK: Numerous points of different configurations lie over the red stars. 

Table \ref{tab:bert-compress} presents the results of different KD-based BERT compression methods. For fair comparison, we do not include other methods described in Section \ref{sec:related_work}, because they either distill different type of knowledge or use different student model structure. Here, we focus on comparing the performance with or without HSK compression given the same student model. We can see that except for the results of a few tasks on the test sets, HSK compression consistently promotes the performance of the baseline methods.

\section{Improving Training Efficiency}
\label{sec:efficiency}

\begin{figure}[t]
\centering
\includegraphics[width=1.0\linewidth]{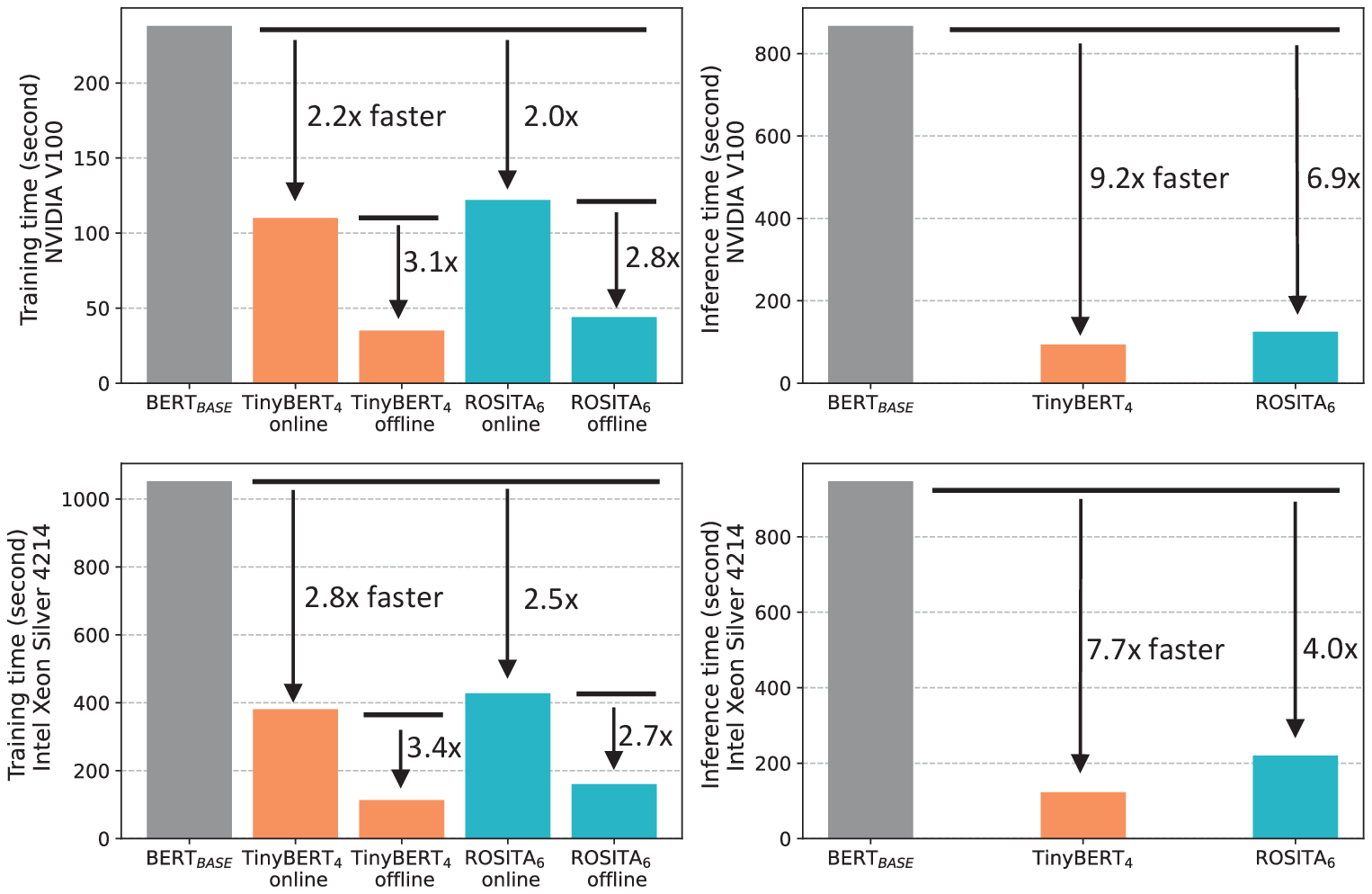}
\caption{Training time (left) and inference time (right) with different devices and models on MNLI. Online means the teacher is loaded during training and offline is the proposed KD paradigm. Please refer to Appendix \ref{sec:efficient_train_set} for the experimental setups.}
\label{fig:infer_time}
\end{figure}
Existing BERT compression methods mostly focus on improving the inference efficiency. However, the teacher model is used to extract features throughout the training process, which suggests that the training efficiency still has room for improvement. As shown in Figure \ref{fig:infer_time}, the compressed models achieve considerable inference speedup, while the increase in training speed is relatively small. Moreover, for students with different sizes or architectures, the teacher should be deployed every time when training a new student. Intuitively, we can run the teacher once and reuse the features for all the students. In this way, we do not need to load the teacher model while training the student, and thereby increasing the training speed. We refer to this strategy as offline HSK distillation \footnote{In the literature \cite{KDSurvey}, ``offline distillation" also means the teacher parameters are fixed during KD, which is different from our definition here.}.

To evaluate the training efficiency of the proposed KD paradigm, we compute the training time on the MNLI dataset. The results are presented in the left plots of Figure \ref{fig:infer_time}. As we can see, offline HSK distillation increases the training speed of the student models, as compared with online distillation. The speedup is consistent for different student models and devices.

Despite the training speedup, however, loading and storing HSK increases the memory consumption. The full set of HSK can take up a large amount of space, especially for the pre-trained language models like BERT. Fortunately, our findings in the previous sections suggest that the student only requires a tiny fraction of HSK.

\begin{table}[t]
\centering
\resizebox{1.0\hsize}{!}{$
\begin{tabular}{@{}l l c c c @{}}
\toprule
$(N^{D}, N^{L}, N^{W})$   &$A^{HSK}$ &Feature Size (GB) &Mag Mask Size (GB)\\ \midrule

(1, 9, 0.1)     
&0.9   &1.0   &2.5         \\

(5, 2, 0.3)   
&3   &3.4   &2.8          \\

(3, 8, 0.2) 
&4.8   &5.4   &6.8      \\

Full $\text{ROSITA}_{6}$    
&896   &1011   &0          \\

\bottomrule
\end{tabular}
$}
\caption{Memory consumption of different $A^{HSK}$ on MNLI training set. The last row is the full set of HSK for $\text{ROSITA}_{6}$. }
\label{tab:hsk_size}
\end{table}

Table \ref{tab:hsk_size} summarizes the actual memory consumption of four configurations with different $A^{HSK}$. As we can see, the full set of HSK for $\text{ROSITA}_{6}$ takes up approximately 1 TB of memory space, which is only applicable to some high-end cloud servers. Compressing the HSK can reduce the size to GB level, which enables training on devices like personal computers. It is worth noticing that storing the dynamic \textbf{Mag Mask} is consuming, which typically accounts for more space than HSK. However, the binary masks can be further compressed using some data compression algorithms.

Based on the above results and analysis, we summarize our paradigm for efficient HSK distillation as: First, the teacher BERT runs on the training data to obtain and store the features of HSK and predictions. This can be done on devices that have sufficient computing and memory resources. Then, according to the target application and device, we decide the student's structure and the amount of HSK to distill. Finally, KD can be performed on a cloud server or directly on the target device.

\section{Related Work}
\label{sec:related_work}
KD is widely studied in BERT compression. In addition to distilling the teacher's predictions as in \citet{HintonKD}, researches have shown that the student's performance can be improved by using the representations from intermediate BERT layers \cite{PKD,Rosita,DynaBERT} and the self-attention distributions \cite{TinyBERT,MobileBERT}. Typically, the knowledge is extensively distilled in a layer-wise manner. To fully utilize BERT's knowledge, some recent work also proposed to combine multiple teacher layers in BERT KD \cite{ALP-KD,BERT-EMD} or KD on Transformer-based NMT models \cite{WhySkip}. In contrast to these studies that attempt to increase the amount knowledge, we study BERT KD from the compression point of view. Similar idea can be found in MiniLMs \cite{MiniLMv2,MiniLM}, which only use the teacher's knowledge to guide the last layer of student. However, they only consider knowledge from the layer dimension, while we investigate the three dimensions of HSK.

We explore a variety of strategies to determine feature importance for each single dimension. This is related to a line of studies called the attribution methods, which attempt to attribute a neural network's prediction to the input features. The attention weights have also been investigated as an attribution method. However, prior work \cite{AttNotExplan,IsAttInterp,OnIndentifiTrans,SelfAttAttr} finds that attention weights usually fail to correlate well with their contributions to the final prediction. This echoes with our finding that the original \textbf{Att} strategy performs poorly in length compression. However, the attention weights may play different roles in attribution and HSK distillation. Whether the findings in attribution are transferable to HSK distillation is still a problem that needs further investigation.

\section{Conclusions and Future Work}
In this paper, we investigate the compression of HSK in BERT KD. We divide the HSK of BERT into three dimensions and explore a range of compression strategies for each single dimension. On this basis, we jointly compress the three dimensions and find that, with a tiny fraction of HSK, the student can achieve the same or even better performance as distilling the full-scale knowledge. Based on this finding, we propose a new paradigm to improve the training efficiency in BERT KD, which does not require loading the teacher model during training. The experiments show that the training speed can be increased by $2.7\times \sim 3.4\times$ for two kinds of student models and two types of CPU and GPU devices.

Most of the compression strategies investigated in this study are heuristic, which still have room for improvement. Therefore, a future direction of our work could be designing more advanced algorithm to search for the most useful HSK in BERT KD. Additionally, since HSK distillation in the pre-training stage is orders of magnitude time-consuming than task-specific distillation, the marginal utility diminishing effect in pre-training distillation is also a problem worth studying.

\section*{Acknowledgments}
This work was supported by National Natural Science Foundation of China (No. 61976207, No. 61906187).

\bibliographystyle{acl_natbib}
\bibliography{acl2021}

\appendix
\section{Architecture of Student Models}
\label{sec:model_archi}
TinyBERT \cite{TinyBERT} rescales the structure of BERT from the number of layers, the dimension of the Transformer layer outputs, and the hidden dimension of feed-forward networks. We use the 4-layer version (14.5M parameters) of TinyBERT that is released by \citet{TinyBERT}.

ROSITA \cite{Rosita} compresses BERT from four structural dimensions, namely the layer, attention heads, the hidden dimension of the feed-forward network and the rank of SVD to compress the embedding matrix. In practice, we scale the four dimensions to construct a 6-layer model $\text{ROSITA}_{6}$ that has approximately the same size as $\text{TinyBERT}_{4}$. $\text{ROSITA}_{6}$ has 6 layers and 2 attention heads, and the FFN dimension and embedding matrix rank are 768 and 128 respectively.

\begin{table*}[t]
\small
\centering
\begin{tabular}{@{}l c c c c c c c@{}}
\toprule
Dataset     &CoLA     &SST-2      &QNLI    &MNLI   &MRPC   &RTE   &STS-B   \\ \midrule

HSK Distillation \\ \midrule
learning rate (constant)       &$5e^{-5}$   &$5e^{-5}$  &$5e^{-5}$  &$5e^{-5}$  &$5e^{-5}$  &$5e^{-5}$   &$5e^{-5}$\\
batch size           &32    &32   &64   &64   &32   &32   &32  \\
max sequence length  &64    &64   &128  &128  &128  &128  &128 \\ 
\# epoch             &30    &10   &5    &5    &20   &20   &20  \\ \midrule

Prediction Distillation \\ \midrule
learning rate (linear decay)       &$2e^{-5}$   &$2e^{-5}$  &$2e^{-5}$  &$5e^{-5}$   &$2e^{-5}$     &$2e^{-5}$     &$2e^{-5}$\\
batch size           &32   &32   &64    &64    &32    &32    &32   \\
max sequence length  &64   &64   &128   &128   &128   &128   &128   \\ 
\# epoch             &5    &5    &5     &5     &5     &5     &5     \\

\bottomrule
\end{tabular}
\caption{Hyperparameters for HSK distillation and prediction distillation.}
\label{tab:hyperparameter}
\end{table*}

\section{Hyperparameters}
\label{sec:hyperparameter}
Following \citet{TinyBERT}, we first distill HSK and then distill the teacher's predictions. The hyperparamers for HSK distillation basically follow \citet{TinyBERT}, except that the training epoch of CoLA is changed from 50 to 30, the training epoch of QNLI is changed from 10 to 5, and the batch size for MNLI and QNLI is changed from 256 to 64. For prediction distillation, we use the linear decaying learning rate schedule. For each model and dataset, we tune the number of epoch (from $\{5, 10\}$) and learning rate (from $\{2e^{-5}, 5e^{-5}\}$) for the baseline method that use the uniform layer-wise strategy $g(l,12)$, and the hyperparameters are used for all the results with compressed HSK. Table \ref{tab:hyperparameter} summarizes the hyperparameters.

\section{Configurations of 3D Compression Strategy}
\label{sec:config}
As described in the paper, for each $A^{HSK}$ we can obtain a number of configurations. Specifically, when we use the $\text{ROSITA}_{6}$ there are 13, 21, 45, 75, 112 configurations for $A^{HSK} = 1\pm10\%, 3\pm10\%, 5\pm10\%, 10\pm10\%, 50\pm10\%$ respectively. We randomly sample subsets of configurations in our experiments, the statistics of which is shown in Table \ref{tab:3d_config}. The configurations that exceed the layer constrain are excluded for $\text{TinyBERT}_{4}$. The detailed configurations for different $A^{HSK}$ are summarized in Table \ref{tab:config_appendix}. 

\section{Experimental Settings for Efficiency Evaluation}
\label{sec:efficient_train_set}
In Figure 9, we show the training and inference time of two models on two devices. The training time is computed as the time to run 500 training steps (i.e, batches of data). When it comes to inference, we run the models on the entire training set and dev set for GPU and CPU respectively. For training, the batch size is set to 64 and 16 for GPU and CPU respectively. For inference, we set the batch size to 128 and 1 for GPU and CPU respectively. The maximum sequence length is 128 for all the settings. For offline distillation, we use the configuration (1, 9, 0.1).

\section{More Experimental Results}
\label{sec:appendix_more_results}

\subsection{Full Results of Depth Compression}
\label{sec:appendix_depth}
\begin{figure*}[t]
\centering
\includegraphics[width=1.0\linewidth]{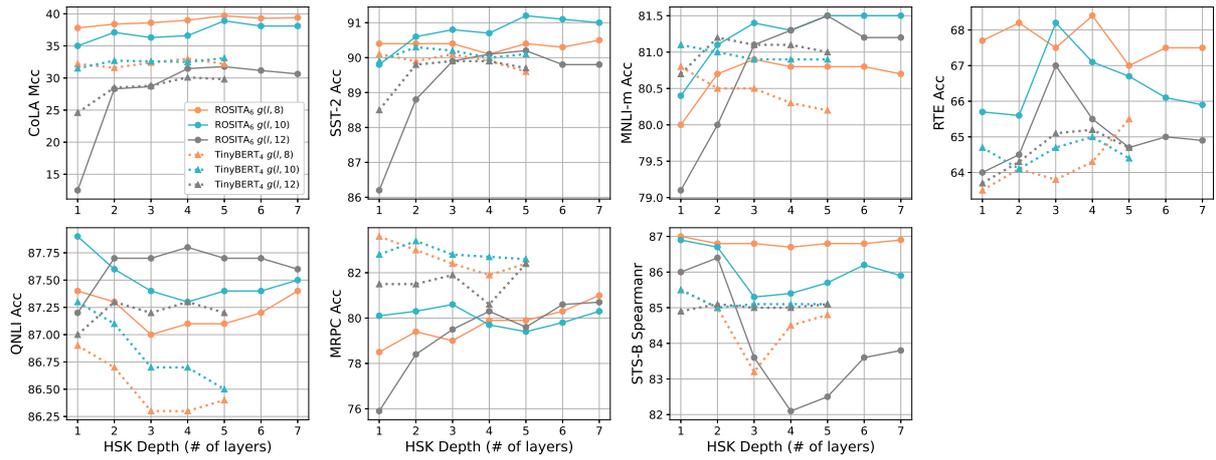}
\caption{Results of depth compression on seven tasks. Each color denotes a layer mapping function.}
\label{fig:depth_appendix}
\end{figure*}

Depth compression results on all seven tasks are presented in Figure \ref{fig:depth_appendix}. Like the results on CoLA, SST-2, QNLI and MNLI, the results on MRPC, RTE and STS-B also suggest that the redesigned mapping functions (i.e., $g(l,8)$ and $g(l,10)$) generally outperforms the original uniform mapping function $g(l,12)$, especially when HSK is compressed to one layer.

\subsection{Full Results of Length Compression}
\label{sec:appendix_length}
\begin{figure*}[t]
\centering
\includegraphics[width=1.0\linewidth]{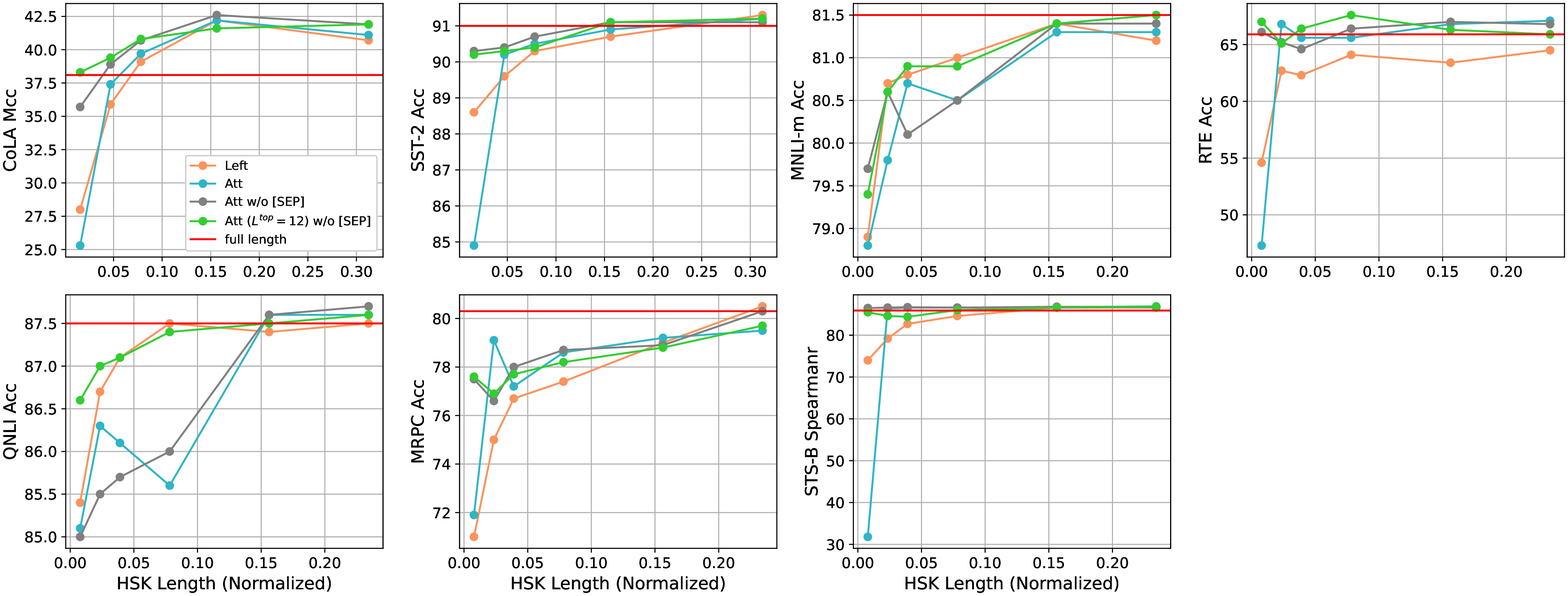}
\caption{Length compression results of $\text{ROSITA}_{6}$ on seven tasks. The horizontal axis represents the compressed HSK length normalized by full length. The left-most points in each plot mean compressing the length to one token.}
\label{fig:length_rosita_appendix}
\end{figure*}

\begin{figure*}[t]
\centering
\includegraphics[width=1.0\linewidth]{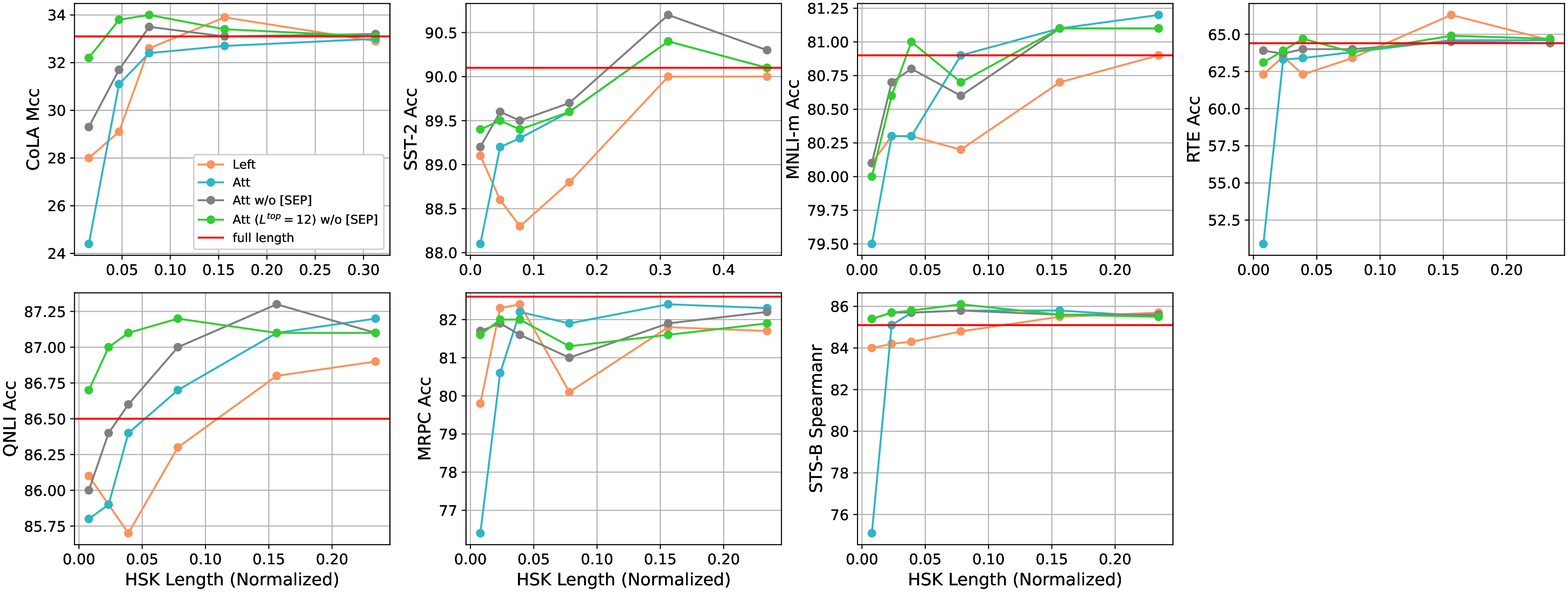}
\caption{Length compression results of $\text{TinyBERT}_{4}$ on seven tasks. The horizontal axis represents the compressed HSK length normalized by full length. The left-most points in each plot mean compressing the length to one token.}
\label{fig:length_tinybert_appendix}
\end{figure*}

Length compression results on all seven tasks are presented in Figure \ref{fig:length_rosita_appendix} and Figure \ref{fig:length_tinybert_appendix}. As we can see, the general trends on MRPC, RTE and STS-B are in accordance with the other four tasks, whose results are discussed in Section \ref{sec:length_compress_analy}. Significant performance drop only occurs when length is compressed to less than 0.05 on MRPC, RTE and STS-B. The strategy base on original attentinon weights (i.e., \textbf{Att}) performs poorly with small HSK length. In comparison, \textbf{Att w/o $[\mathrm{SEP}]$} and \textbf{Att ($L^{top}=12$) w/o $[\mathrm{SEP}]$} reduce the performance drop caused by length compression.

\begin{figure*}[t]
\centering
\includegraphics[width=1.0\linewidth]{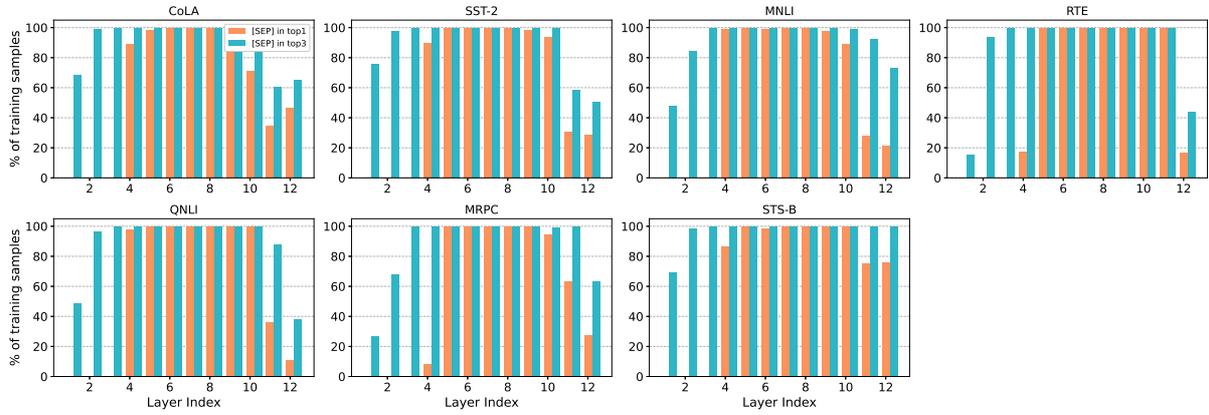}
\caption{The proportion of the data samples in which $[\mathrm{SEP}]$ is among the top1 and top3 attended tokens. We present the results over the 12 layers of the $\mathrm{BERT}_{\mathrm{BASE}}$ fine-tuned on seven tasks.}
\label{fig:sep_rank_appendix}
\end{figure*}

Figure \ref{fig:sep_rank_appendix} shows the proportion of data samples where $[\mathrm{SEP}]$ is the top1 and top3 most attended token. We can see that for most data samples, $[\mathrm{SEP}]$ is among the top3 tokens and frequently appears as the top1 from the $4^{th} \sim 10^{th}$ layers. This pattern is consistent across the seven tasks.

\subsection{Full Results of Width Compression}
\label{sec:appendix_width}
\begin{figure*}[t]
\centering
\includegraphics[width=1.0\linewidth]{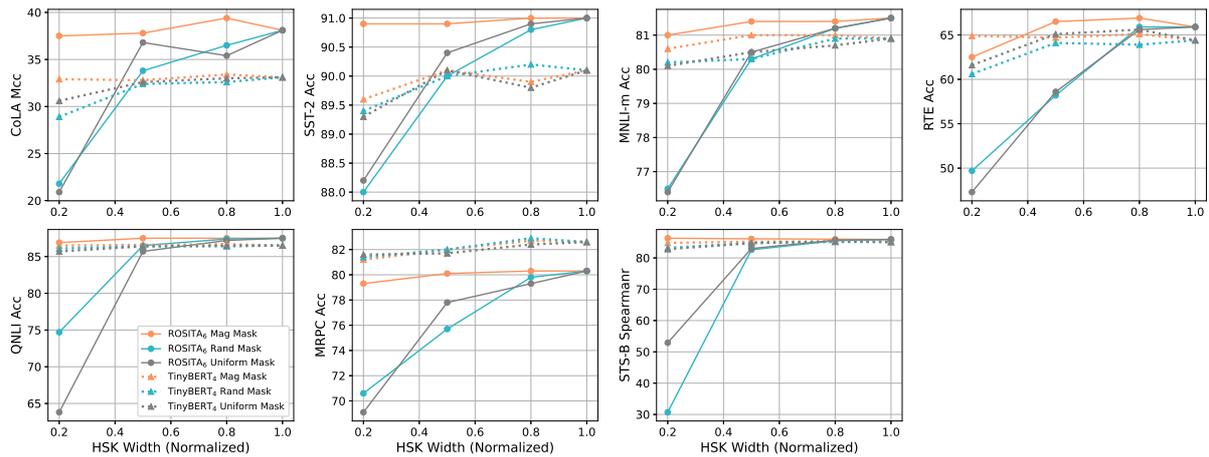}
\caption{Results of width compression with different masking strategies on seven tasks.}
\label{fig:width_appendix}
\end{figure*}

Figure \ref{fig:width_appendix} shows the full results of width compression on all seven tasks. We can see that the gap between compression strategies is larger for $\text{ROSITA}_{6}$, as compared with $\text{TinyBERT}_{4}$. Among the three strategies, \textbf{Mag Mask} clearly outperforms \textbf{Rand Mask} and \textbf{Uniform Mask}.

\begin{table}[t]
\centering
\resizebox{1.0\hsize}{!}{$
\begin{tabular}{@{}l l c c c @{}}
\toprule
1$\pm10\%$    &3$\pm5\%$     &5$\pm5\%$       &10$\pm5\%$    &50$\pm5\%$\\ \midrule

(2, 1, 0.5) & \bf(6, 1, 0.5) & (1, 13, 0.4) & \bf(7, 3, 0.5) & \bf(6, 10, 0.8)  \\
(5, 2, 0.1) & \bf(6, 5, 0.1) & \bf(6, 2, 0.4) & (1, 49, 0.2) & (2, 36, 0.7)  \\
(1, 9, 0.1) & (1, 29, 0.1) & (5, 5, 0.2) & (3, 8, 0.4) & (3, 25, 0.7)  \\
(1, 2, 0.5) & (1, 10, 0.3) & (2, 24, 0.1) & (3, 5, 0.7) & (2, 27, 0.9)  \\
(1, 5, 0.2) & (1, 5, 0.6) & (1, 25, 0.2) & (1, 25, 0.4) & \bf(6, 9, 0.9)  \\
(1, 1, 1.0) & (2, 15, 0.1) & (1, 6, 0.8) & (2, 7, 0.7) & \bf(6, 12, 0.7)  \\
(3, 3, 0.1) & (5, 2, 0.3) & \bf(7, 7, 0.1) & (3, 16, 0.2) & \bf(6, 27, 0.3)  \\
(3, 1, 0.3) & (5, 1, 0.6) & \bf(6, 8, 0.1) & \bf(7, 14, 0.1) & \bf(7, 24, 0.3)  \\
(1, 10, 0.1) & (5, 6, 0.1) & (4, 4, 0.3) & (2, 8, 0.6) & (4, 18, 0.7)  \\
(2, 5, 0.1) & (3, 5, 0.2) & (2, 6, 0.4) & (5, 3, 0.7) & (3, 21, 0.8)  \\
(1, 1, 0.9) & (1, 30, 0.1) & (2, 3, 0.8) & (3, 7, 0.5) & (5, 10, 1.0)  \\
(5, 1, 0.2) & (5, 3, 0.2) & (3, 8, 0.2) & (4, 8, 0.3) & \bf(7, 23, 0.3)  \\
(1, 3, 0.3) & (1, 6, 0.5) & (4, 12, 0.1) & (5, 7, 0.3) & (5, 25, 0.4)  \\
 & (1, 15, 0.2) & (1, 7, 0.7) & \bf(6, 8, 0.2) & (4, 26, 0.5)  \\
 & (2, 3, 0.5) & (1, 49, 0.1) & (5, 5, 0.4) & \bf(6, 17, 0.5)  \\
 & (1, 3, 1.0) & (2, 4, 0.6) & \bf(6, 2, 0.8) & (2, 26, 1.0)  \\
 & (2, 5, 0.3) & (5, 10, 0.1) & (5, 4, 0.5) & \bf(6, 8, 1.0)  \\
 & (3, 10, 0.1) & (1, 5, 1.0) & (4, 5, 0.5) & (4, 12, 1.0)  \\
 & (3, 2, 0.5) & (3, 4, 0.4) & (1, 13, 0.8) & (2, 43, 0.6)  \\
 & (3, 1, 1.0) & (1, 10, 0.5) & (1, 32, 0.3) & (5, 16, 0.6)  \\
 &  & (2, 13, 0.2) & (3, 33, 0.1) & (4, 25, 0.5)  \\
 &  & (3, 2, 0.8) & (3, 34, 0.1) &   \\
 &  & (4, 3, 0.4) & (4, 25, 0.1) &   \\
 &  & (1, 8, 0.6) & \bf(6, 4, 0.4) &   \\
 &  & (5, 2, 0.5) & (5, 10, 0.2) &   \\
 &  & (2, 5, 0.5) & (2, 26, 0.2) &   \\
 &  & \bf(6, 1, 0.8) & \bf(7, 15, 0.1) &   \\
 &  & (4, 6, 0.2) & (1, 11, 0.9) &   \\
 &  & (1, 16, 0.3) & (4, 6, 0.4) &   \\
 &  & (4, 13, 0.1) & (2, 6, 0.8) &   \\
 &  & \bf(6, 4, 0.2) & (2, 10, 0.5) &   \\
 &  &  & (1, 10, 1.0) &   \\
 &  &  & (4, 26, 0.1) &   \\
 &  &  & (5, 2, 1.0) &   \\
 &  &  & (3, 4, 0.8) &   \\
 &  &  & (4, 3, 0.8) &   \\

\bottomrule
\end{tabular}
$}
\caption{Configurations $(N^{D}, N^{L}, N^{W})$ of 3D HSK compression for different $A^{HSK}$. The configurations in bold font are not used for $\text{TinyBERT}_{4}$.}
\label{tab:config_appendix}
\end{table}

\end{document}